# Towards Computing Inferences from English News Headlines

Elizabeth Jasmi George[1][0000-0001-6012-5364] and Radhika Mamidi[1][0000-0003-0171-0816]

[1] LTRC, International Institute of Information Technology, Hyderabad, India
`elizabeth.george@research.iiit.ac.in, radhika.mamidi@iiit.ac.in`

**Abstract.** Newspapers are a popular form of written discourse, read by many people, thanks to the novelty of the information provided by the news content in it. A headline is the most widely read part of any newspaper due to its appearance in a bigger font and sometimes in colour print. In this paper, we suggest and implement a method for computing inferences from English news headlines, excluding the information from the context in which the headlines appear. This method attempts to generate the possible assumptions a reader formulates in mind upon reading a fresh headline. The generated inferences could be useful for assessing the impact of the news headline on readers including children. The understandability of the current state of social affairs depends greatly on the assimilation of the headlines. As the inferences that are independent of the context depend mainly on the syntax of the headline, dependency trees of headlines are used in this approach, to find the syntactical structure of the headlines and to compute inferences out of them.

**Keywords:** Computing Inferences, Presuppositions, Conventional implicatures, Pragmatics, News Discourse, News Headline.

## 1. Introduction

The headline of a news report appears at the top of the news report and is often printed in a bigger font and some times in bright colour. The marketability of a news story depends to a great extent on the ability of the headline to attract readers. A headline generally tries to summarise the content of the news story, with a strong intention of communicating the context to the reader. Headlines also try to attract the attention of the newsreaders, prompting them to read on through the news story. Headline functions as a number of speech acts. It urges, warns and informs the reader [11]. This work views headline as a potential source of rich information capable of generating multiple inferences relevant to the current social state making it worthy of adding to the general knowledge.

This work was done as a part of building a system for children to learn about current affairs in a simpler way. In this work, we consider headline as a standalone unit of discourse, without any context or supporting background information and compute the inferences that arise from the headline alone. Our experiment attempts to compute inferences based on syntactical triggers.

This paper focuses on inferences, in particular presuppositions and conventional implicatures which are independent of context and omit conversational implicature which requires context information to formulate. The number of triggers used in this experiment is limited and the results include negatives in some cases.

### 1.1. Presupposition and Conventional Implicature

According to Levinson [15] presupposition is used to describe any kind of background assumption against which an action, theory, expression or utterance makes sense or is rational and conventional implicatures are non-truth-conditional inferences that are not derived from superordinate pragmatic principles like the maxims but are simply attached by convention to particular lexical items or expressions.

According to Fromkin et al. [6], presuppositions are implicit assumptions about the world, required to make an utterance meaningful or appropriate. Not unlike lexical presuppositions conventional implicatures are associated with specific words and result in additional conveyed meanings when those words are used, according to Yule [21]. Presuppositions are denoted by '>>' and conventional implicatures are denoted by '≈'. For an utterance "*The King of France is Wise.*" there can be a presupposition that >> *There is a present king of France*. For an utterance "*Amelia is a Toddler but she is quiet.*" there can be a conventional implicature that ≈ *Toddlers are not usually quiet* [3].

As an example, upon reading a headline '*Schaeuble says British were 'deceived' in Brexit campaign*', a reader may make the following inferences. (i). *Schaeuble exists*. (ii). *Schaeuble said something*. (iii). *Schaeuble believes that the British were 'deceived' in Brexit campaign*. (iv). *Brexit campaign happened*. (v). *Brexit can have campaign* (vi). *The British government was deceived in the Brexit campaign* (vii). *The British citizens were deceived in the Brexit campaign*. The inferences (vi) and (vii) which are conversational implicatures need more contextual information along with the headline under consideration to support them. So generating inferences like (vi) and (vii) is not attempted in this work and we try to generate inferences similar to those stated from (i) to (v).

### 1.2. Related Work

Cianflone et al. [1] have introduced the novel task of predicting adverbial presupposition triggers, and this paper explores the scope of computing presupposition statements from the syntax structure provided by dependency trees of news headlines. The approach used in that paper uses deep learning while this paper demonstrates a rule-Based approach. The RTE task [22] dataset consisted of *text(t)-hypothesis(h)* pairs with the task of judging for each pair whether *t* entails *h*. In this work, we attempt to generate hypotheses for news headlines rather than judging whether a hypothesis is correct. Burger and Ferro [23] attempted to generate a large corpus of textual entailment pairs from the lead paragraph and headline of a news article. To the best of our knowledge, ours is the first work towards computing presuppositions and conventional implicatures from English news headlines.

### 1.3. Linguistic Definitions and Characteristics of Headlines

According to Dor [4], headlines are "the negotiators between stories and readers" and they have four functions of summarising, highlighting, attracting and selecting. The headline together with the lead or the opening paragraph summarises a news story. Gattani [8] identifies three broad macro headline functions. (i). The informative headline, which gives a good idea about the topic of the news story. (ii). The indicative headline, which addresses what happened in the news story. (iii). Eye catcher headline, which does not inform about the content of the news story but is designed to entice people to read the story. The greater the mental effort required for processing a headline, the less relevant it becomes [4]. While reading a headline the reader should be able to construct assumptions, either based on what can be perceived in their immediate environment or on the basis of assumptions already stored in their memory. The relevance of a headline is directly proportional to the amount of contextual effects and inversely proportional to the cognitive processing effort required to recover these effects [18].

Headlines are characterised by the density of the information present in them and they have the syntactic characteristics of telegraphic speech. They also contain bold expressions, polarisation, exaggerations and provocative wording [14]. While processing headlines, more information should be expected from a shorter span of words. The grammatical rules for proper English sentences would be frequently violated either for filling more information in the short space available or for promoting the curiosity of the reader. News headlines use a special language called 'block language', a name first coined by Straumann [9]. Block language has a structure different from the normal clause or sentence structure but it often conveys a complete message. This language usually consists of lexical items lower than sentences.

### 1.4. Relevance of this Work

This work computes inferences from headlines. The inferences generated can be fed to a learning system which grades the impact created by the headline, based on sensitivity, child-Friendliness, clarity and various other parameters as required. It is advantageous to evaluate the impact because an ordinary reader naturally reads through the headlines in the newspaper before starting to read the whole news articles. The understandability of the headline contributes towards the ease of understanding of the news story that follows it.

## 2. Data

The dataset used in this work is comprised of around 350 headlines collected manually from different news websites [25-27] about four popular events which appeared continuously in news reports for a time span of a few months. The topics selected for including in the dataset are 'Brexit', 'Disputes over the South China Sea', 'Syrian refugee crisis' and 'Pyeongchang Winter Olympics'. In the dataset, the headlines were arranged in chronological order to facilitate their use in studying the gradual evolution of the headlines, assuming that the reader has already read the previous headlines for the same news item. The timestamp associated with headlines in the dataset is not used in the present work, though it might be useful for future developments to evaluate how headlines evolve as the news on that topic progresses in course of time and how readers understand them based on their awareness of the previous headlines on the same topic.

### 2.1. Format of Data

The data used as input for computing inferences using our rule-Based system are in the format: Headline [source: News source Timestamp]. A subset of the same dataset is used for collecting human inferences for evaluation purpose. Some Examples of headline data is given below.
*U.S. vows new North Korea sanctions ahead of Olympics face-off* [source: Reuters February 07, 2018 06:39 PM IST]
*Schaeuble says British were "deceived" in Brexit campaign* [source: Reuters June 23, 2017 07:18 PM IST]

## 3. Proposed Method

In this work, it is assumed that only the headline is available to the reader for understanding the topic of the news and that the reader is completely ignorant of the previous happenings under the same topic of news. The inferences of headlines are computed based on some logical conclusions attained, rooted in certain grammatical relations present in the headline. Rusu et al. [20] suggest subject-predicate-object triplet extraction from sentences which motivated this work. In the case of a news headline, the participants are the composer of the headline, who is the speaker and the common person reading the headline, who is the addressee. For computing inferences, we begin with the extraction of nouns and verbs. The algorithm is outlined below.

---
**Algorithm 1.** Computing Inferences from a news headline:

---
1: Extract one headline from the dataset and preprocess it by removing the punctuations.
2: Annotate the headline with POS tags for all tokens in it, using Stanford CoreNLP [16].
3: Get all the verbs in the headline by comparing the POS tags of the tokens against the regular expression 'V.+'.
4: Get corresponding dependencies for all the verbs of the headline, using Stanford CoreNLP annotated with 'depparse'. Refer section 3.1
5: Get all nouns and pronouns from the headline by comparing the POS tags of the tokens against the regular expression 'N.+|P.+'.
6: Generate explicit inferences from headline using Stanford OpenIE [7]
7: Generate more inferences using the rule-Based system under section 3.2 based on grammatical relations held between tokens in the headline.

---

In the algorithm, we start with dependency parsing the headline, thus obtaining the verbs occurring in the headline with their dependencies. We get the headline tagged with POS tagger from Stanford and then extract the list of nouns and list of verbs in the headline. The verbs are also lemmatised to get the base form of the verbs present in the headline. The lemmatised form is used when a different form of the verb other than the tense form in which it appears in the headline, is required for a changed tense form in the computed inferences. A few rule-Based approaches are implemented to get inferences from the headline. Stanford openIE [7] gives inferences which are directly stated in the headline. The headline "How the company kept out 'subversives'" gives the inference "company kept out 'subversives' " by openIE [7]. More inferences assumed from the syntactical structure of the headline are generated by the rule-Based system.

### 3.1. Extracting Dependencies

The Stanford dependencies are binary grammatical relations held between a 'governor' and a 'dependent' as specified in the Stanford dependencies manual [17], which provides documentation for the set of dependencies defined for English. The dependencies obtained from the Stanford CoreNLP dependency parser [16] are generated as a dependency tree which contains dependencies as tuples like those in the examples given below for the headline '*Rescue rules by Bank of England will divide Britain*'.

(i). {'dep': 'nmod', 'governor': 2, 'governorGloss': 'rules', 'dependent': 4, 'dependentGloss': 'Bank'}
(ii). {'dep': 'case', 'governor': 6, 'governorGloss': 'England', 'dependent': 5, 'dependentGloss': 'of'}
(iii). {'dep': 'nmod', 'governor': 4, 'governorGloss': 'Bank', 'dependent': 6, 'dependentGloss': 'England'}
(iv). {'dep': 'aux', 'governor': 8, 'governorGloss': 'divide', 'dependent': 7, 'dependentGloss': 'will'}
(v). {'dep': 'dobj', 'governor': 8, 'governorGloss': 'divide', 'dependent': 9, 'dependentGloss': 'Britain'}

### 3.2. Rule-Based System for Inference Generation

In this work, we use a rule-Based system that is comprised of rules based on commonly occurring syntactical patterns. These patterns are modelled as inference triggers. Inference generation logic for an associated inference trigger is configured as a rule. Multiple iterations are performed on the dependency relations to generate inferences. Node JS tense conjugator [24] is used to find the required tense form of the verb to be attached in the computed inferences.

Since this work demonstrates the use of syntax structures to generate inferences using only a few triggers in the scope of inference triggers, the addition of more known triggers like iterative — *anymore, return, another time, to come back, restore, repeat* etc. change of state verbs — *stopped, began, continued, start, finish, carry on, cease, leave, enter, come, go, arrive* etc. Factive verbs — *regrets, aware, realise, know, be sorry that, be proud that, be indifferent that, be glad that, be sad that* etc. Verbs of judging—*accuse, criticise, blame, apologise, forgive, condemn, impeach* etc. which humans are better at making inferences upon should be included for more ac-curate results, by elaborating the rules using string comparison of the verb under consideration with these above-mentioned triggers. The Current set of inference triggers and rules used in computing inferences from headlines are listed below. The set of rules can be extended with more patterns to improve the quality of inferences.

**Presence of a Future Tense Verb.** Presence of a future tense verb in the headline could suggest that we can infer that the event described by the noun is yet to happen.

If dependent is 'aux'(auxiliary) and 'dependentGloss' is the string 'will' then iterate once again through the dependencies to find a dependent 'dobj'(direct object) which is the noun phrase which is the (accusative) object of the verb where the 'governorGloss' of both dependency relations match.

Eg: "*Russian state television will not broadcast Olympics without national team.*" can have an inference >>"*Olympics is not yet broadcast* ".

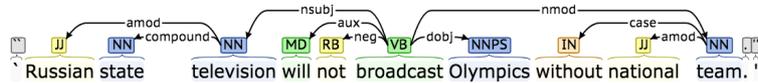

**Fig. 1.** Dependency structure for the headline '*Russian state television will not broadcast Olympics without national team*'.

---

**Algorithm 2.** Computing Inferences Based on the Presence of a Future Tense in a Headline:

---

1: Consider VD as the set of verbs in the headline with their dependencies, obtained from parser
2: **for each** dependency tuple D **in** VD
3: **if** 'dep' of D is= 'aux' **and** 'dependentGloss' of D is = 'will' **then**
4: **for** each dependency tuple ND **in** VD
5: **if** 'dep' of ND is = 'dobj' **and** 'governorGloss' of ND is = 'governorGloss' of D **then**
6: **output** 'dependentGloss' of ND
7: **output** "is not yet"
8: **output** past tense of ('governorGloss' of D)

---

**Presence of the Conjunction 'but'.** Presence of the conjunction 'but' could suggest that we can infer that the subject was expected to undergo 'negation' of that which is mentioned in the part of the headline after the conjunction 'but'.

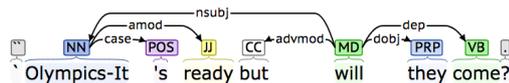

**Fig. 2.** Dependency structure for the headline '*Olympics-It's ready but will they come?*'

Eg: "*Olympics-It's ready but will they come?*" can have a inference >>" *Olympics - being ready was expecting coming*".

---

**Algorithm 3.** Computing Inferences Based on the Presence of Conjunction 'but' in a Headline:

---

1: Consider VD as the set of verbs in the headline with their dependencies, obtained from parser
2: **for each** dependency tuple D **in** VD
3: **if** 'dep' of D is = 'conj:but' **then**
4: **output** "being "
5: **output** 'governorGloss' of D
6: **output** " was [not] expecting "
7: **output** Gerund of ('dependentGloss' of D )

---

**Presence of 'again' in a Clause with a Verb.** Presence of 'again' as an adverbial modifier in a clause with a verb could suggest that we can infer that the event described by the noun has already happened.

Eg: "*Norway regulator again rejects "Donut" fish farm volume plan.*" can have an inference >>"*Norway regulator has rejected "Donut" fish farm volume plan before*".

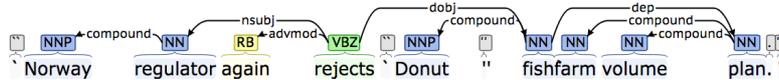

**Fig. 3.** Dependency structure for the headline '*Norway regulator again rejects "Donut" fish farm volume plan*'

---

**Algorithm 4.** Computing Inferences of a Headline Which has Presence of 'again' in a Clause with a Verb:

---

1: Consider VD as the set of verbs in the headline with their dependencies, obtained from parser
2: Consider N is the set of all nouns in the headline
3: **for each** dependency tuple D in VD
4: **if** 'dep' of D is = 'advmod' **and** 'dependentGloss' of D is 'again' **or if** any noun **in** N is 'dependentGloss' with 'dep' of D = 'nsubj' **then**
5: **for each** dependency tuple ND **in** VD
6: **if** 'dep' of ND is = 'nsubj' **and** 'governorGloss' of ND is = 'governorGloss' of D **then**
7: **output** 'dependentGloss' of ND
8: **output** past tense of ('governorGloss' of D) "before"

---

Get all the nouns in the headline and iterate through them until the 'dependentGloss' of a tuple is a noun in the headline and the dependent is 'nsubj'(nominal subject) that is a noun phrase which is the syntactic subject of a clause or if dependency relation is 'advmod'(adverb modifier). Then if the 'governorGloss' is 'again' follow from step 5 of Algorithm 4.

**Presence of 'further' as an Adverb.** Presence of 'further' as an adverb could suggest that we can infer that now it is already in the state described by the 'noun' related to the verb modified by the adverb 'further'.
Eg: "*UK economy to slow further.*" can have an inference >>"E*conomy is already slow*".

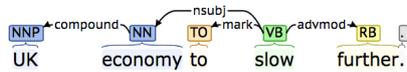

**Fig. 4.** Dependency structure for the headline '*UK economy to slow further*'

**Presence of a 'noun compound'.** Presence of noun compound like 'Brexit campaign' could suggest that we may infer that 'Brexit' that is the first part 'N1' of the noun compound can be /can have a 'campaign', that is the second part 'N2' of the noun compound. The problem of computing semantic relation of the nouns N1 and N2 in the noun compound is not dealt with in this experiment. Only common sense assimilation that "N1 can be N2" or "N1 can have N2" is generated.
Eg: "*Russia's Olympic ban strengthens Putin's reelection hand.*" can have an inference >>"*Olympic can be /can have ban*".

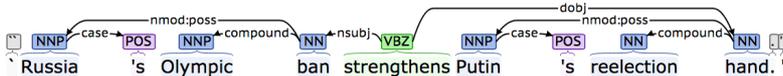

**Fig. 5.** Dependency structure for the headline '*Russia's Olympic ban strengthens Putin's reelection hand*'

**Presence of a 'verb' in Past Tense.** If the 'verb' is in the past tense in a headline it could suggest that we can infer that, the event has already happened.
Eg: "*The dude released this video before he went on a killing spree*" can have an inference >>"*dude has released this video*".

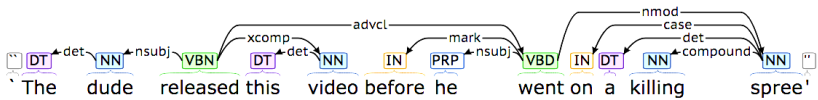

**Fig. 6.** Dependency structure for the headline '*The dude released this video before he went on a killing spree*'

**Presence of Nominal Modifier 'of'.** If there is a nominal modifier 'of' then it could suggest that we can infer that the dependent 'has' governor.
Eg: "*Bank of England plans rescue.*" can have an inference >> "*England has Bank*".

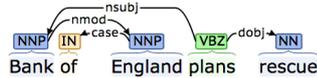

**Fig. 7.** Dependency structure for the headline '*Bank of England plans rescue*'

## 4. Results and Discussion

The unavailability of annotated inferences makes the comparison and evaluations difficult for this task. The inferences generated with the system are compared with manually annotated inferences for 100 randomly collected headlines. Annotators are two research scholars doing research in Linguistics and fluent in English. They did the annotation of the subset of the dataset for evaluation manually, based on the annotation guidelines provided to them (see section 6.2 of the Appendix). Annotation guidelines with explanatory examples for the inference triggers mentioned in section 3.2 were given to the annotators and they were asked to look for the surface structure of the headline in general and use human judgement in making inferences.

No upper limit on the number of generated human inferences was imposed. 11.8% of the inferences generated by the annotators were of the existential types, such as those beginning with a clause like "there exists". The inference triggers other than the existential ones are occurring less in headlines compared to normal discourse, due to the peculiarity of block language used.

**Table 1.** Accuracy and Generated Percentage of Inferences Computed

| Inference Trigger | Percentage of Accurate Inferences | Percentage of Inaccurate Inferences | Percentage of Missing Inferences |
|---|---|---|---|
| But | 69.3 | 0 | 30.7 |
| Again | 82.7 | 8.3 | 9 |
| Further | 94 | 6 | 0 |
| Future Tense | 93 | 3 | 4 |
| Noun Compound | 54.4 | 40.2 | 5.4 |

The percentages of computed inferences for some inference triggers used in this experiment is given in Table 1. For a headline '*Britain takes step towards Brexit with repeal bill*' our system generates the following inferences (i). *Britain takes step* (ii). *Britain takes step towards Brexit* (iii). *Britain takes step with repeal bill* (iv). *repeal can be/can have bill* (v). *Brexit has step*.

**Table 2.** Comparison of manually annotated inferences with computed inferences for a headline

| Headline | Manually Annotated Inferences | Computed Inferences | Percentage of Correct Inferences | Percentage of Incorrect results |
|---|---|---|---|---|
| IOC extends North Korea deadline for Pyeongchang games | 1. IOC has power to extend deadline<br><br>2. North Korea has deadline<br><br>3. Deadline can be extended<br><br>4. There exists North Korea<br><br>5. There exists Pyeonchang games | 1. Korea can have deadline<br><br>2. Pyeongchang has games<br><br>3. Games has deadline | 40% | 0% |
| Olympics:Medals at Winter-Olympics through years | 1. There exists Winter Olympics<br><br>2. Olympics has medals<br><br>3. Olympics had been happening through years<br><br>4. There exists medals in years Olympics was conducted | 1. Winter can have olympics<br><br>2. Olympics has medals<br><br>3. years had medals | 75% | 0% |
| Schaeuble Says British were "deceived" in Brexit campaign | 1. Schaeuble exists<br><br>2. Schaeuble believes that the British were "deceived" in Brexit campaign<br><br>3. Brexit can have campaign<br><br>4.Schaeuble said something.<br><br>5. Schaeuble believes that the British were 'deceived' in Brexit campaign.<br><br>6. Brexit campaign happened. | 1. Schaeuble Says British were "deceived"<br><br>2. Brexit can be/ can have campaign<br><br>3. campaign has deceived | 16.7% | 33% |

Table 2 shows the comparison results of manually annotated inferences with the computed inferences for the three headlines in the first column and gives the percentage of correct computed inferences and percentage of incorrect results out of the computed inferences for those headlines. For example for the last headline — "*Schaeuble Says British were "deceived" in Brexit campaign*" only one of the manually annotated inferences — "*Brexit can be/can have campaign*" is computed by our Rule-Based system thus making the percentage of correct computed inferences to be 16.7%, and out of the three computed inferences "*campaign has deceived*" is wrong and thus the percentage of incorrect results in the computed inferences is 33%.

## 5. Conclusions and Future Work

In this work, we considered headline as a stand-alone unit of text without attaching any information from the context in which it appeared in a news report. Based on the observation that the presence of certain words and tense conditions can trigger inferences from a headline, we tried to generate inferences based on a set of rules, formulated based on certain grammatical relations present in the headline. In future, the rule set could be expanded to include more observations and complex rules to compute more inferences. These inferences can be used to measure the impact and sensitivity of a headline mainly for checking the appropriateness when used in a platform designed for children. This experiment was more of an attempt towards computing inferences from the headline and the results are not complete due to the limited proportion of rules implemented compared to the large list of cases generating presuppositions and conventional implicatures. This approach of applying logic on the syntactic structure to generate inferences stand different from alternative approaches using deep learning techniques because of the lesser data, time and compute requirement.

**Acknowledgements.** We would like to thank Dr.Monojit Choudhury, Microsoft Research- Bangalore, for suggesting this topic of research as part of the Computational Socio-pragmatics course he taught at IIIT-H. We would also like to thank all the anonymous reviewers for carefully reading through our manuscript and offering valuable suggestions.

**Appendix**

## 6. Annotation Guidelines

### 6.1. Purpose of Annotation

This Annotation task targets to provide the possible presuppositions for a news headline. Presuppositions can be any background assumption against which the headline makes sense or is rational. Presuppositions are denoted by a '>>' symbol. A sentence and its negative counterpart share the same set of presuppositions, so the headline "Karnataka CM meets prime minister Narendra Modi " will have the following pre-supposition >> "Narendra Modi is the prime minister" which is true for the statement "Karnataka CM meets prime minister Narendra Modi" as well as its negative counterpart "Karnataka CM does not meet prime minister Narendra Modi ".

**6.2. Guidelines for Annotating Presuppositions**

For annotating, look for presupposition triggers, which are the linguistic items that are particular words or some aspects of the surface structure of the headline in general, which generates presuppositions. The following are some presupposition triggers with examples.

**Definite Descriptions.**
*Example*. Hunterston B: Pictures show cracks in Ayrshire nuclear reactor
>> There exists cracks in Ayrshire nuclear reactor.

**Factive Verbs.** Factive verbs like *regrets, aware, realize, know, be sorry that, be proud that, be indifferent that, be glad that, be sad that* etc.
*Example*. Corbyn 'regrets' Labour MPs' resignations
>> Labour MPs resigned.

**Implicative Verbs.** Implicative verbs like *manage, remember, bother, get, dare, care, venture, condescend, happen, be careful, have the misfortune, have the sense, take the time, take the trouble, take the opportunity* etc.
*Example*. How Russia Managed to Destroy Saudi Arabia ?
>> Russia destroyed Saudi Arabia.

**Change of State Verbs.** Change of state verbs like *stopped, began, continued, start, finish, carry on, cease, leave, enter, come, go, arrive* etc.
*Examples*. (i). Britain continued to struggle with Brexit
>> Britain was struggling with Brexit.
(ii). China has stopped stockpiling metals.
>> China had been stockpiling metals.

**Iteratives.** Iteratives like *again, anymore, return, another time, to come back, restore, repeat, for the nth time* etc.
*Examples*. (i). HTC in talks with Micromax, Lava and Karbonn to return to Indian market
>> Micromax, Lava and Karbonn had been in Indian market previously.
(ii). BoE's Carney says will reassess outlook when there is Brexit clarity
>> Outlook has been assessed before.

**Verbs of Judging.** Verbs of judging like *accuse, criticise, blame, apologize, forgive, condemn, impeach* etc.
*Examples*. (i). Trump blames financial market 'disruption' on Democrats
>> Trump thinks that financial market disruption is bad.
(ii). Amnesty criticises Hungary over treatment of migrants

>> Amity thinks that Hungary was treating migrants bad.

**Temporal Clauses.** Temporal clauses like *before, while, after, when, during, whenever* etc.
*Example*. Britons were endlessly lied to during Brexit campaign
>> There was a Brexit campaign.

**Cleft Sentences.** Cleft sentences like i. What he wanted to buy was a Fiat, ii. It is Jaime for whom we are looking, iii. All we want is peace etc.
*Example*. It is Jaime for whom we are looking
>> We are looking for someone.

**Implicit Clefts with Stressed Constituents.** Implicit clefts with stressed constituents like capital letters, or bold type, or underlined type can give rise to presuppositions.

**Comparisons and Contrasts.** Comparisons and contrasts like *too, back, in return* etc. can give rise to presuppositions.
*Example*. Russia is a better negotiator than Italy
>> Italy is a negotiator.

**Non-restrictive Relative Clauses.**
*Example*. John, who passed the test, was elated.
>> John passed the test.

**Counterfactual Conditionals.**
*Example*. If I had a guarantee, then I'd love them
>> I don't have a guarantee.

**Questions.**
*Example*. What's missing from your low carb breakfast?
>> Something is missing from your low carb breakfast.
   Similarly *who* can be replaced by *someone*, *where* by *somewhere*, *how* by *somehow* to generate presuppositions. *Yes/No* questions will generally have vacuous presuppositions.
*Example*. Are you living with mild or moderate depression?
>> Either you are living with mild or moderate depression or you are not.

**More than Two Words in Quotes.** More than two words in quotes can give a presupposition that something is said. News headlines sometimes have quotes to emphasize words. So it may not be an utterance always. So we assume that more than 2 words in quotes mean something is said.
*Example*. Merkel says May's Brexit proposals "not the breakthrough".
>> Merkel says "not the breakthrough".

**Future Tense Verb.** Presence of future tense verb in the headline can create a presupposition that the event described in the noun has not happened yet.
*Example.* Russian state television will not broadcast Olympics without national team
>> Olympics is not yet broadcast by Russian state television.

**The Conjunction 'but' Suggest a Contrast.**
*Example.* Olympics-It's ready but will they come?
>> Being ready was expecting them to come.

**Gender-Specific Statements.**
*Example.* New Zealand Prime Minister Jacinda Ardern gives birth to first child.
>> Jacinda Arden is a female.

Since the headlines use tricky language to attract readers, human intuition while listing the presuppositions is required. Format of the annotation is to write presuppositions preceding with a '>>' following the headline, and after writing all presuppositions for a headline, ending it with a '||' with one presupposition statement in a line. Presuppositions should be expressed as simple sentences in simple English.